\pdfoutput=1

\documentclass[11pt]{article}

\usepackage[]{acl}

\usepackage{times}
\usepackage{latexsym}
\usepackage{booktabs}

\usepackage[T1]{fontenc}

\usepackage[utf8]{inputenc}
\usepackage{graphicx}
\usepackage{xspace}

\usepackage{microtype}
\usepackage[compact]{titlesec}

\newcommand{\placeholder}{\textsc{PrOf}\xspace}
\newcommand{\repository}{\url{https://github.com/dnozza/profanity-obfuscation}\xspace}

\title{The State of Profanity Obfuscation in Natural Language Processing}

\author{Debora Nozza, Dirk Hovy \\
  Bocconi University \\
  Via Sarfatti 25 \\
  Milan, Italy \\
  \texttt{\{debora.nozza,dirk.hovy\}@unibocconi.it} }

\begin{document}
\maketitle
\begin{abstract}
Work on hate speech has made the consideration of rude and harmful examples in scientific publications inevitable. This raises various problems, such as whether or not to obscure profanities. While science must accurately disclose what it does, the unwarranted spread of hate speech is harmful to readers, and increases its internet frequency. While maintaining publications' professional appearance, obfuscating profanities make it challenging to evaluate the content, especially for non-native speakers.
Surveying 150 ACL papers, we discovered that obfuscation is usually employed for English but not other languages, and even so quite uneven.
We discuss the problems with obfuscation and suggest a multilingual community resource called \placeholder that has a Python module to standardize profanity obfuscation processes. We believe \placeholder can help scientific publication policies to make hate speech work accessible and comparable, irrespective of language.
\end{abstract}

\noindent
\textit{\textbf{Warning}: this paper contains unobfuscated examples some readers may find offensive.}

\section{Introduction}
A major downside of unsavory research subjects is that they still need to be investigated and reported, especially if we want to improve matters. Hateful language poses this challenge in natural language processing. To detect, classify, and mitigate it, we first need to collect and annotate it. Setting aside the ethical conundrum of subjecting annotators to hateful language \cite{kennedy_2018,vidgen-etal-2019-challenges}, reporting on it presents researchers with a range of challenges.

On the one hand, science should unflinchingly report on its subject matter, no matter how unpleasant \cite{miso}. On the other hand, if that subject matter is language, then reporting on it is almost equivalent to producing it. This issue presents two problems: 1) proliferation and 2) audience framing.

Context should disambiguate whether a word is used in its intended meaning or as a \textit{meta function} of talking \textit{about} the word \cite{jakobson-2010} without using its original intent. However, written content that is freely accessible on the internet might appear in unexpected contexts: e.g., as training data for language models \cite{NEURIPS2020_1457c0d6}.

Open access also means that it is not clear who will read a given text. Scientific readers might assume the meta function and discount hateful language, but there is no guarantee as to who other readers may be. Disclaimers can help frame this problem and give the reader a choice. However, \textbf{readers who do not want to read unpleasant examples should not be excluded from conducting research in hate speech detection}. Disregarding their personal feelings about an offensive term seems cruel and insensitive at best, especially if they are members of targeted groups o abuse victims (see Table~\ref{tab:examples}).

A compromise solution is \textbf{obfuscation}, where one or more letters in an offensive term are replaced with stars or other symbols. This approach preserves the word shape and allows interested readers to reconstruct the original word without allowing it to proliferate it or forcing it upon readers.

\begin{table}[]
\centering
\small
\begin{tabular}{p{\dimexpr0.5\columnwidth\tabcolsep-\arrayrulewidth\relax}|
                p{\dimexpr0.34\columnwidth\tabcolsep-\arrayrulewidth\relax}              } 
                \toprule
A political h*mo? I am not listening to a fairy gay f*ggot [...] & \cite{zhu-bhat-2021-generate}   \\ \midrule
suck a pig d*ck c*nt                                                                       & \cite{botelho-etal-2021-deciphering}  \\   \midrule

Bruh im tired of n*ggas [...] & \cite{shvets-etal-2021-targets} \\ \midrule

Someone should r*pe her & \cite{guest-etal-2021-expert} \\
                                                             
                                                                                           \bottomrule
\end{tabular}
\caption{Examples originally reported as unobfuscated in research papers. Here we obfuscate them.}
\label{tab:examples}
\end{table}

However, based on our survey of 150 NLP papers, there are several issues with obfuscation:
\begin{itemize}
    \item \textbf{Obfuscated words are often not discernible}, especially for non-native (English) speakers. Profanities are not taught in school, and we cannot expect people learning English to know and recognize them when characters are hidden \cite{Dewaele_2004}. This kind of language changes more readily than more formal language - even older native speakers might not recognize novel slurs. Moreover, it is basically impossible to search for obfuscated words without guessing their meaning (not to mention the impact on the search history).\footnote{None of the authors of this paper are native speakers of English and all have faced this issue.}
    \item Authors have \textbf{many choices when obfuscating} words, e.g., obfuscate only vowels, keep or obfuscate only the first letter, etc. As a result, profanities can be obfuscated in a wide variety of ways, making their interpretation even harder. E.g., \textit{cunt}, \textit{c*nt}, \textit{c**t},  \textit{c***}.
    \item There is \textbf{no clear definition} of what is a profanity and what should be obfuscated, especially if a word has other, more neutral, meanings. E.g., \textit{retarded} or \textit{r*tarded}.
    \item Profanities in \textbf{languages other than English} tend not to be obfuscated.
\end{itemize}

This prompts one simple question: \textit{How can we use profanity obfuscation in scientific publication}?
Among the 150 *ACL publications from 2021 reporting profanities, a number of solutions emerge. Even if standardization has been made in specific venues\footnote{\url{https://www.workshopononlineabuse.com/resources-and-policies/reporting-examples}}, these solutions are not consistent.

As outlined above, obfuscation leads to several unintended problems, predominantly for non-native speakers.
As the NLP community grows, an increasing number of readers face this conundrum.
\textit{N**ger} is easy enough to guess, but \textit{p*cker} is difficult without advanced knowledge. Or, if you are a native speaker of English, consider Danish \textit{p*rker}, German \textit{F*tz*}, or Italian \textit{bo***ino}. Now try googling them.
Moreover, many slurs and insults are culture-specific. For example, without knowledge of the history of racism in America, it is almost impossible to even guess at the meaning of \textit{c**n}. 
This issue is related to the bias towards work on English in NLP \cite{bender-friedman-2018-data}.

\paragraph{Contributions} We surveyed profanity reporting in 150 scientific publications. Based on our findings, we propose \placeholder (Profanity ObFuscation), a multi-lingual resource to help researchers converge on common procedures for profanity obfuscation. \placeholder will permit researchers to report profanities in scientific publications while ensuring formal appearance, readability, and accessibility.

\section{Do we need a framework for profanity obfuscation?}
Research in hate speech inevitably needs to face the use of profanities in language. These taboo words are known to be perceived negatively \cite{psy1,psy2}, leading to heightened states of emotional arousal \cite{jay2008recalling}, and potentially causing vicarious trauma \cite{vidgen-etal-2019-challenges}.
As scholars publishing on open access platforms, we need on the one hand to protect readers from this content and, on the other hand,  to report the message in its unexpurgated entirety \cite{miso} because euphemisms and generic descriptors cannot convey the hostility. Obfuscation, if not ideal, is the best compromise to deal with the conundrum of hateful language in scientific literature: \textbf{obfuscated words should be discernible to allow accessibility and replicability}. How can a researcher test the same examples if it is not possible to discern the text?
At the same time, while the content deciphering is left to the reader, the conveyed emotion is still negative \cite{stout2015examination}.


\section{Methodology and Results}
\label{sec:method}
We surveyed the ACL Anthology for proceedings of *ACL conferences that took place in 2021 and a workshop specifically focused on abuse detection, the Workshop on Online Abuse and Harms (WOAH). 
We searched this data for occurrences of ``*'' and ``\#'', used to obfuscate profanities.\footnote{When the proceedings did not include all papers in a single file, we searched in each paper the mention of \textit{abuse}/\textit{hate}/\textit{offensive}/\textit{toxic} in the title or abstract.} For each paper that included one or more profanities, we noted whether or not the authors notify the use of offensive language and which languages the authors considered in their offensive examples. Each conference's profanity count is listed in Table \ref{table:stats}.

\begin{table}[]
\centering
\small
\begin{tabular}{lr}

\textbf{Proceedings} & \begin{tabular}[x]{@{}c@{}}\textbf{\# obfuscated}\\\textbf{profanities}\end{tabular} \\
\toprule
\textbf{ACL 2021}       & 67                    \\
\textbf{EACL 2021}           & 15                                           \\
\textbf{EMNLP 2021}          & 11                                                \\
\textbf{WOAH 2021}           & 57         \\     
\bottomrule
\end{tabular}
\caption{Statistics of papers using obfuscation\label{table:stats}.}
\end{table}

\subsection{Current practice}

Several approaches can be used for obfuscation. To minimize the possibility of offending readers, it is possible to completely obfuscate the word or maintain only the first letter. E.g. \textit{fuck} would result in ``\textit{****}'' or ``\textit{f***}'' or ``\textit{f*}''. However, this practice makes the words (almost) impossible to decipher.

The most common practice is to obfuscate vowels. For example, \textit{fucking} would become "\textit{f*cking}" or "\textit{fuck*ng}" or "\textit{f*ck*ng}". This hypothetically makes the meaning intelligible by suppressing the fewest number of characters. 


Summarizing the scientific publications in the *ACL community, the current practice is: (1) Obfuscation is always performed via the "*" symbol, (2) there is no shared practice of which letters to obfuscate, and (3) there are different sensibility levels when choosing which words to obfuscate, especially in languages other than English.

\subsection{Considerations regarding obfuscation}

Lack of a uniform profanity obfuscation in scientific articles affects readability and accessibility.

\paragraph{Lack of * use consistency} 
Obfuscation is highly \textbf{inconsistent across different papers}. Some authors remove the first vowel \cite{xu-etal-2021-detoxifying,chuang-etal-2021-mitigating,luccioni-viviano-2021-whats,xu-etal-2021-bot,elsherief-etal-2021-latent}, others obfuscate two letters (e.g., \textit{f**king}) \cite{qian-etal-2021-lifelong,sheng-etal-2021-societal,turcan-etal-2021-emotion,bhat-etal-2021-say-yes}, or obfuscate the first letter (e.g., \textit{*ucking}) \cite{kang-hovy-2021-style}, or other customized choices \cite{ousidhoum-etal-2021-probing,gros-etal-2021-r,sawhney-etal-2021-multitask,mishra-etal-2021-modeling-users,baheti-etal-2021-just}. Some choices may lead to sentences that are not understandable, e.g., ``\textit{All you n* and s*}'' \cite{du-etal-2021-self}.
A more important issue is the \textbf{lack of consistency within the same paper}, further compounding the confusion around profanity obfuscation practices. For example, in \newcite{sheng-etal-2021-nice,mostafazadeh-davani-etal-2021-improving}, the authors obfuscate almost all letters for some words but few for others (e.g.,  \textit{f***} and \textit{a**hole}), and \newcite{salawu-etal-2021-large} use both \textit{p*ssy} and \textit{pu**y}. If the same word is obfuscated differently, though, readers may think they are actually different words, maybe unknown (e.g., \textit{putty}).

\paragraph{Word obfuscation choices}
Another problem is the choice of \textit{whether} to obfuscate a word. Some authors choose to also obfuscate words that are not vulgar per se, such as \textit{dumb} or \textit{queer} \cite{caselli-etal-2021-hatebert,rottger-etal-2021-hatecheck}, but that may be offensive in a specific context, i.e., when used as an insult. Again, we found a \textbf{lack of consistency in obfuscation choices in the same paper}. This means that some authors decide to obfuscate some words (e.g., \textit{ni***r}) but not others (e.g., \textit{whore}) \cite{cheng-etal-2021-mitigating,vidgen-etal-2021-learning,bagga-etal-2021-kidding,laugier-etal-2021-civil,zhou-etal-2021-challenges}.

\paragraph{Typos} 
We also observed typos in obfuscated words. This can generate confusion for readers who might misinterpret these mistakes as unknown profanities. For example, we found \textit{b*itch} \cite{bertaglia-etal-2021-abusive} and \textit{wh*ore} \cite{kirk-etal-2021-memes}.

\paragraph{No obfuscation} 
A number of papers reported profanities without any form of obfuscation \cite{shvets-etal-2021-targets,fortuna-etal-2021-min,hahn-etal-2021-modeling,nozza-2021-exposing,zampieri-etal-2020-semeval,sen-etal-2021-counterfactually,chiril-etal-2021-nice-wife,an-etal-2021-predicting-anti,cercas-curry-etal-2021-convabuse,xie-etal-2021-models,dale-etal-2021-text,mehrabi-etal-2021-lawyers,leonardelli-etal-2021-agreeing,zhu-bhat-2021-generate,botelho-etal-2021-deciphering,guest-etal-2021-expert,hede-etal-2021-toxicity}. 
Some of these works study other languages in addition to English. Since the scientific research is English-centric, the authors potentially found the profanities in other languages less hurtful \cite{gonzalez1972anxiety,gleason_2003,CHRISTIANSON201773}.

A possible solution for not using obfuscation in hate speech detection is to select examples that do not contain profanities \cite{niraula-etal-2021-offensive}.
However, we argue that scholars are responsible for reporting hate speech as severe as it is, no matter how unpleasant \cite{miso}.
Note that offensive language can occur in other non-hateful contexts as well \cite{malmasi2018challenges}.


\paragraph{Multimodality} 
A challenging issue is profanities in images containing text, such as memes or artifact figures. While the same procedures outlined above could be applied, a solution is that (1) images created by the authors should conform to the standards, while (2) they can report images from the internet in their original form, but with a disclaimer on the paper's first page. We observe this procedure in several publications \cite{zia-etal-2021-racist,kougia-pavlopoulos-2021-multimodal,qian-etal-2021-lifelong,elsherief-etal-2021-latent,baheti-etal-2021-just}. There are still exceptions where artefact figures report unobfuscated profanities \cite{an-etal-2021-predicting-anti,bucur-etal-2021-exploratory,zhou-etal-2021-challenges}.

\subsection{Considerations regarding disclaimers}

Less than 20\% of NLP papers use disclaimers of offensive content. However, the community needs to reach a behavioral standard. Knowing where and how disclaimers should be placed is important to ensure every reader is aware of the use of offensive examples in the paper.
The papers including disclaimers applied very different practices. Disclaimers are placed (1) before the abstract \cite{xu-etal-2021-bot,mehrabi-etal-2021-lawyers}, (2) after the abstract \cite{cercas-curry-etal-2021-convabuse}, (3) as a footnote on the first page \cite{nozza-2021-exposing,zampieri-etal-2020-semeval}, or (4) under the table of offensive examples \cite{kang-hovy-2021-style,elsherief-etal-2021-latent}. Most papers warning users of offensive language do not use any form of obfuscation in the paper. 
\textbf{We recommend authors add an italicized disclaimer at the end of the abstract to signal that a paper includes offensive terms}. This practice should be implemented even when profanities are obfuscated.

\section{\placeholder}
We propose \placeholder, a multi-lingual community resource for the obfuscation of profanities in scientific publications. It allows for the uninterrupted reading of papers with profanities while allowing non-native speakers to look up words and definitions if they desire. For the definition, we use the multi-lingual BabelNet \cite{NAVIGLI2012217}.
\placeholder consists of a table reporting:\\
1) the unobfuscated profanity (e.g., \textit{fuck})\\
2) first-vowel obfuscation (e.g., \textit{f*ck})\\
3) the language (e.g., \textit{English})\\
4) the part-of-speech (POS) tag (e.g., \textit{NOUN})\\
5) the BabelNet multi-lingual synset (e.g., \url{https://babelnet.org/synset?id=bn:00006453n&lang=EN}) or other resources if the synset does not exist.

We suggest the obfuscation practice of removing the first vowel. For compound words, we obfuscate the first vowel of the element with an offensive meaning (e.g., \textit{femin*zis}).

We extend \placeholder to other languages with the help of native speakers, reaching a total of 203 profanities: 50 in English, 44 in French, 19 in German, 42 in Italian, and 48 in Spanish. Details about \placeholder construction are given in Appendix \ref{app:construction}.

We understand that our work is currently limited to the profanities of the languages we speak, and the set of profanities we cover. However, we argue that: with respect to the past, we cover all the profanities reported in \textasciitilde3000 published papers; in the future, \placeholder is meant to be a community research that grows along the research that will use it.


\paragraph{Using \placeholder} We release \placeholder as a Python package\footnote{\repository} and web application. The package automatically obfuscates profanities starting from a string or a text file, and can reveal profanities from their obfuscated versions (see Appendix \ref{sec:app}).



\section{Related Work}
Profanities have been investigated in NLP for discovering how to automatically filter them or how to prevent their obfuscation. These issues can be solved straightforwardly with a forbidden word list. However, preparing this list is difficult, as people are constantly creating new forms to avoid filtering via dictionary lookups, such as \textit{\$h!t}, \textit{sh1t}, or \textit{s.h.i.t}. I.e, introducing spacing or punctuation between letters, swapping or removing characters, and 0--9 substitutions.
Approaches to automatically filter variations of vulgar words are based on string matching techniques \cite{yoon2010smart,ghauth2015text}. The research on de-obfuscation of profanities is much larger. This is due to NLP tools' need to access the content of a sentence. Several studies \cite{mishra-etal-2018-author,mishra-etal-2018-neural,eger-etal-2019-text,mehdad-tetreault-2016-characters} showed that obfuscated words are often ignored or treated as out-of-vocabulary impacting tasks like sentiment analysis or hate speech detection. Methods range from sequence alignment algorithms used in genomics \cite{10.1145/3032963} to word embeddings \cite{lee2018abusive,renwickdetection}.
Our work differs from this literature in that we focus on scientific publications, not on social media. 
In this setting, the use of a dictionary is feasible.


\section{Conclusion}
Our work highlights the lack of obfuscation standards for reporting profanities in scientific publications. Prevailing practice allows for dangerous procedures and restricted access.
We introduce \placeholder, a resource to standardize profanity obfuscation in scientific publications. \placeholder allows researchers to prevent offending readers while ensuring that information is readable and accessible.
We plan to expand to more languages. As researchers add new words featured in their papers, \placeholder will grow along with the number of publications.


\section*{Ethical Consideration}
We consider as profanities words that have highly offensive or vulgar connotations. We acknowledge that readers may have different sensibilities with respect to profanities. Obscene words depend on different factors, such as culture, social or religious background, and more. 
Consequently, some words may be disturbing for a number of people, and should be obfuscated, while other readers may not have any issue with reading them. Moreover, we should consider that there is typically a hierarchy of offense, whereby some words are more severe than others; for example,  \textit{f*ck} is often socially accepted while the \textit{n-word} usually is not \cite{sap-etal-2019-risk}.

\section*{Acknowledgements}

This project has partially received funding from the European Research Council (ERC) under the European Union’s Horizon 2020 research and innovation program (grant agreement No.\ 949944, INTEGRATOR), and by Fondazione Cariplo (grant No. 2020-4288, MONICA).
Debora Nozza and Dirk Hovy are members of the MilaNLP group and the Data and Marketing Insights Unit of the Bocconi Institute for Data Science and Analysis.

\bibliography{anthology,custom}
\bibliographystyle{acl_natbib}

\appendix

\section{\placeholder construction}
\label{app:construction}

\placeholder construction starts with the list of English profanities surveyed from recent proceedings of *ACL conferences (see Section \ref{sec:method}). This starting list comprises 50 entries, of which 37 unique terms and 4 unique POS tags (ADJ, ADV, NOUN, VERB). Note that profanities can be associated with different POS tags (e.g., \textit{f*ck} can be a noun, a verb, and an adverb). Table \ref{tab:topwords} lists the most common English profanities and their associated obfuscated version.
We used these 50 English profanities as a seed for creating German, Italian, French, and Spanish \placeholder. Given a profanity, we retrieve all its associated concepts in another language exploiting BabelNet. Note that each language is characterized by a different number of profanities that can be associated with a target group (e.g., women). Using BabelNet instead of a translation tool enables us to retrieve all these terms instead of just one exact translation.
The limitation of this approach is that the number of retrieved concepts starting from one term is very high and not all relevant. For example, some terms can be used to refer to profane acts in some contexts, but their main meaning is non-profane (e.g., \textit{avvitare} (\textit{screw}) is a word that can also be used for referring to the act of having sexual intercourse). In other cases, the profanities related concepts in BabelNet are still in English or are literally translated, resulting in nonsense terms in the target language (e.g., \textit{piece of tail}\footnote{\url{https://www.urbandictionary.com/define.php?term=Piece\%20of\%20Tail}} is literally translated in Italian with the non-existing idiom \textit{pezzo di coda}). For this reason, we filter the retrieved related concepts with a hurtful lexicon (HurtLex) \cite{bassignana2018hurtlex}. Finally, we asked native speakers to validate the resulting filtered list of terms by removing terms that were not unambiguous profanities. We also permit native speakers to include additional profanities if they felt some popular ones were missing, on average they added 4 profanities. The final \placeholder resource comprises 203 profanities: 50 in English, 44 in French, 19 in German, 42 in Italian, and 48 in Spanish.

\begin{table}[]
\small
\centering
\begin{tabular}{llr}

\textbf{obfuscated word} & \textbf{count} \\ 
\toprule
f*ck    & 20 \\
n*gga   & 14 \\
b*tch   & 13 \\
f*cking & 9  \\
f*g     & 8  \\
n*gger  & 8  \\
sh*t    & 8  \\
sl*t    & 7 \\   
\bottomrule
\end{tabular}

\caption{Most common obfuscated profanities in 2021 *ACL proceedings with their counts.\label{tab:topwords}}
\end{table}

\section{Data Statement}
We follow \newcite{bender-friedman-2018-data} on providing a Data Statement for the proposed \placeholder resource. 

Language-specific profanities have been validated by native speakers of each language (French, German, Italian, and Spanish). The annotators are in the age group of 25-35 and have experience in computational linguistics. The data we share is not sensitive to personal information, as it does not contain information about individuals. 

\section{Python package}
\label{sec:app}

\begin{figure}[ht]
\centering
\includegraphics[width=1\columnwidth]{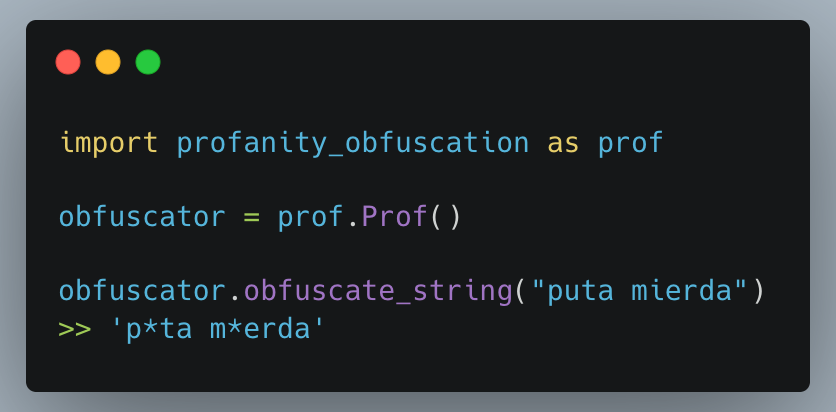}
\caption{Usage examples of the Python package for obfuscating text from a string.}
\label{fig:code1}
\end{figure}

\begin{figure}[]
\centering
\includegraphics[width=1\columnwidth]{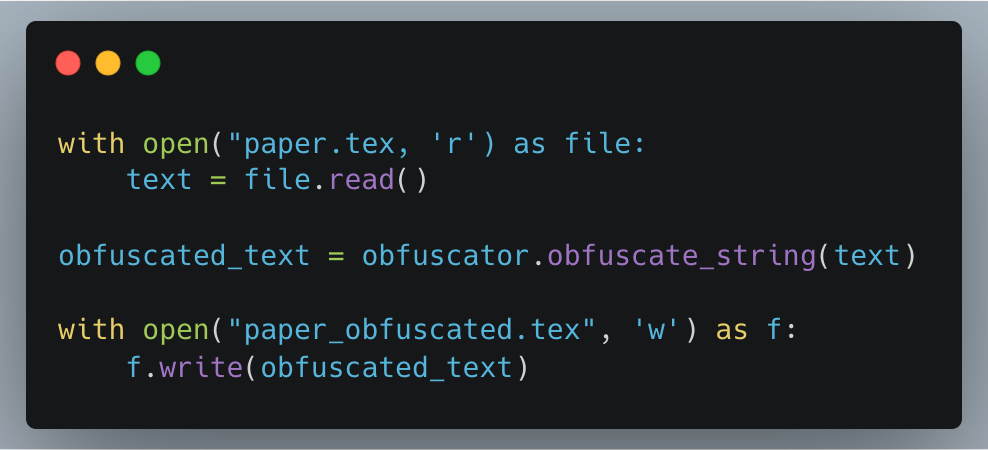}
\caption{Usage examples of the Python package for obfuscating text from a file.}
\label{fig:code2}
\end{figure}

\begin{figure}[]
\centering
\includegraphics[width=1\columnwidth]{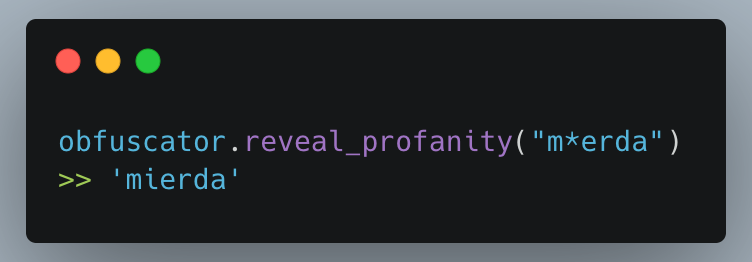}
\caption{Usage examples of the Python package for revealing obfuscated profanities.}
\label{fig:code3}
\end{figure}

\begin{figure*}[]
\centering
\includegraphics[width=0.8\textwidth]{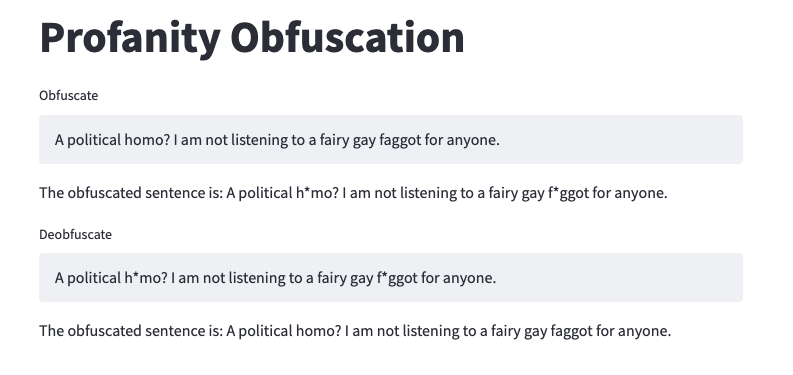}
\caption{Usage examples of the web-app for obfuscating and revealing obfuscated profanities using the example in Table \ref{tab:examples}.}
\label{fig:code4}
\end{figure*}

We released \placeholder as a Python package under the MIT license. We report some code snippets for demonstrating how the library can be used to obfuscate a profanity from a string (Figure \ref{fig:code1}) or from a text file, like a \LaTeX source (Figure \ref{fig:code2}). Figure \ref{fig:code3} shows how our library can be used for revealing a profanity from its obfuscated version.
Finally, Figure \ref{fig:code4} demonstrates the use of \placeholder as a web application for obfuscating and de-obfuscating profanities.

\end{document}